\documentclass[10pt,twocolumn,letterpaper]{article}

\usepackage{cvpr}              %
\usepackage{pdfpages}
\usepackage[dvipsnames]{xcolor}

\newcommand{\mparagraph}[1]{\vspace{0.5mm}\noindent{\textbf{#1.}\hspace{0.5mm}}}

\usepackage{import} %
\graphicspath{{figs/}}

\usepackage[ruled,noend]{algorithm2e} %

\SetAlFnt{\small}
\SetAlCapFnt{\small}
\SetAlCapNameFnt{\small}
\SetAlCapHSkip{0pt}
\IncMargin{-\parindent}
\usepackage{ifpdf}
\usepackage{graphicx}
\ifpdf
  
\else
\fi
\usepackage{color}
\definecolor{cyan}{cmyk}{1,0,0,0}
\definecolor{darkgreen}{rgb}{0,0.5,0}
\definecolor{orange}{rgb}{1,0.5,0}
\definecolor{magenta}{cmyk}{0,1,0,0}
\definecolor{darkyellow}{cmyk}{0,0,0.75,0}
\definecolor{gray}{rgb}{0.8,0.8,0.8}

\usepackage{amsmath} %

\usepackage[toc,page]{appendix} %
\usepackage{wrapfig} %
\usepackage{comment}
\usepackage{bm}
\usepackage{cuted} %
\usepackage{lipsum} %
\usepackage{mathtools} %
\usepackage{algorithmicx}
\usepackage{algpseudocode}
\usepackage{nicefrac}

\usepackage{gensymb}
\usepackage{float}
\usepackage{soul}
\usepackage{array}
\usepackage{multirow}
\usepackage{hhline}
\usepackage{setspace}
\usepackage{nicefrac}

\algdef{SE}[DOWHILE]{Do}{doWhile}{\algorithmicdo}[1]{\algorithmicwhile\ #1}%

\makeatletter
\renewcommand{\ALG@beginalgorithmic}{\small}
\makeatother

\newcommand{\DELETE}[1]{} %
\newcommand{\IGNORE}[1]{}
\usepackage{datenumber}
\usepackage{calc}
\usepackage[mmddyyyy]{datetime}

\newcounter{datetoday}
\newcounter{diffyears}
\newcounter{diffmonths}
\newcounter{diffdays}

\newcommand{\difftoday}[3]{%
      \setmydatenumber{datetoday}{\the\year}{\the\month}{\the\day}%
      \setmydatenumber{diffdays}{#1}{#2}{#3}%
      \addtocounter{diffdays}{-\thedatetoday}%
      \ifnum\value{diffdays}>0
        \def\diffbefore{}%
        \def\diffafter{left}%
      \else
        \def\diffbefore{}%
        \def\diffafter{ago}%
        \setcounter{diffdays}{-\value{diffdays}}%
      \fi
      \setcounter{diffyears}{\value{diffdays}/365}%
      \setcounter{diffdays}{\value{diffdays}-365*\value{diffyears}}%
      \setcounter{diffmonths}{\value{diffdays}/30}%
      \setcounter{diffdays}{\value{diffdays}-30*\value{diffmonths}}%
      \diffbefore
      \ifnum\value{diffyears}=0
      \else
        \ifnum\value{diffyears}>1
            \thediffyears\space years,
        \else
            \thediffyears\space year,
        \fi
      \fi
      \ifnum\value{diffmonths}=0
      \else
        \ifnum\value{diffmonths}>1
            \thediffmonths\space months
        \else
            \thediffmonths\space month
        \fi
      \fi
      \ifnum\value{diffdays}=0
      \else
        \ifnum\value{diffdays}>1
            \thediffdays\space days
        \else
            \thediffdays\space day
        \fi
      \fi
      \diffafter
}

\definecolor{cvprblue}{rgb}{0.21,0.49,0.74}
\usepackage[pagebackref,breaklinks,colorlinks,citecolor=cvprblue]{hyperref}

\def\paperID{17785} %
\def\confName{CVPR\xspace}
\def\confYear{2026\xspace}

\title{Revisiting Pose Sensitivity in Splat-based Computed Tomography\\under Sparse-view Reconstruction}

\author{Kiseok Choi \hspace{1cm} Hyeongjun Cho \hspace{1cm} Inchul Kim \hspace{1cm} Min H. Kim\\[2mm]
KAIST\\[2mm]
{\tt\small \{kschoi;hjcho;ickim;mhkim\}@vclab.kaist.ac.kr}
}

\begin{document}
\maketitle
\begin{abstract}
X-ray computed tomography (CT) reconstructs volumetric representations of objects from projection images obtained by transmitting X-rays through a target. 
Recent splat-based tomography, which represents a volume as a continuous distribution of 3D Gaussians, has demonstrated both high reconstruction quality and fast convergence in cone-beam sparse-view CT. 
However, when deployed in real CT systems with limited and non-uniform view distributions, we observe distinctive streak and strip artifacts that are far more pronounced than in conventional reconstruction methods. 
Through detailed analysis, we show that these artifacts primarily originate from pose inaccuracies in the acquisition geometry rather than from view sparsity itself. 
We revisit pose sensitivity in the splatting formulation and derive a stable gradient-based framework that jointly refines geometric parameters during reconstruction. 
Our study not only identifies how pose perturbations propagate through the differentiable projection operator but also reveals why splat-based CT is particularly vulnerable to geometric misalignment. 
The resulting formulation remains lightweight and easily integrable into existing pipelines while substantially improving reconstruction fidelity under real-world sparse-view conditions.

\end{abstract}

\begin{figure}
	\centering
	\vspace{-5mm}	
	\includegraphics[width=1.0\linewidth]{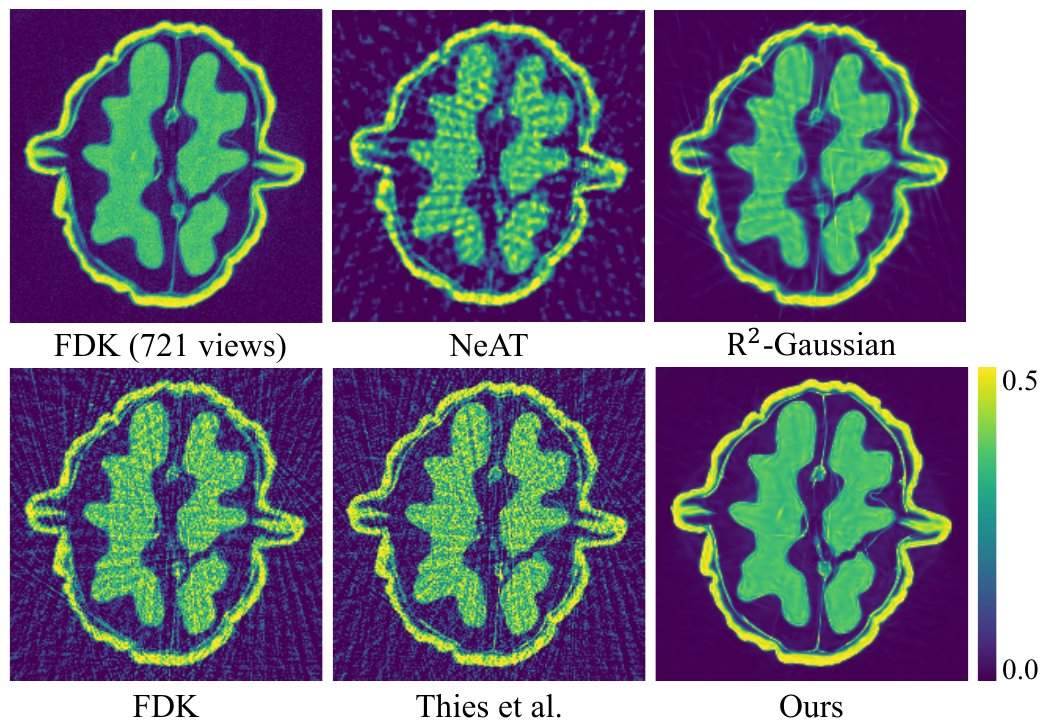}%
	\vspace{-3mm}
	\caption{\label{fig:teaser}
Reconstruction of a real walnut from cone-beam CT using FDK~\cite{feldkamp1984practical} (721 views), joint reconstruction–calibration methods (NeAT~\cite{neat2022}, Thies et al.~\cite{thies2024gradient}), R$^2$-Gaussian~\cite{r2_gaussian}, and our method (others use 75 sparse views). FDK suffers from severe radial artifacts and blur. NeAT reduces noise but introduces directional streaks. Thies et al. improve edges but leave stripe artifacts. R$^2$-Gaussian shows needle artifacts from pose errors. Our self-calibrating splat-based method suppresses these artifacts while preserving fine details, highlighting the importance of geometric calibration in real sparse-view CT.	
}
	\vspace{-4mm}
\end{figure}

\section{Introduction}
\label{sec:intro}
X-ray computed tomography (CT) reconstructs the internal structure of an object by measuring the attenuation of X-rays transmitted through it~\cite{kak2001principles}. 
It plays an essential role in both medical diagnostics and industrial inspection, providing non-destructive access to volumetric structures. 
Among various system geometries such as parallel-beam, fan-beam, and cone-beam configurations~\cite{hsieh2003computed}, cone-beam CT (CBCT) has become the most widely adopted due to its ability to reconstruct high-quality 3D volumes with fewer projections. 
The Feldkamp–Davis–Kress (FDK) algorithm~\cite{feldkamp1984practical} remains a standard solution for fast reconstruction, while iterative optimization techniques~\cite{sart_andersen,asd_pocs} improve quality at the cost of computation.

Recent advances in differentiable rendering and implicit representation have transformed CT reconstruction into an optimization problem on continuous volumetric fields. 
Methods based on neural implicit functions~\cite{naf_zha,sax_nerf,neat2022} and, more recently, splat-based representations~\cite{r2_gaussian} have demonstrated superior performance in sparse-view reconstruction. 
By modeling the attenuation field as a set of anisotropic 3D Gaussians, splat-based tomography achieves both high-quality reconstruction and rapid convergence. 
However, despite its success on synthetic data, we observe that applying this approach to real CT acquisitions introduces distinctive streak and strip artifacts (Figure~\ref{fig:walnut_synthetic_real_reconstruction}(a)). 
This discrepancy indicates that the degradation is not solely due to data sparsity, but arises from underlying geometric inconsistencies in the acquisition process.

Through detailed analysis, we identify geometric calibration errors—particularly inaccuracies in camera poses—as the dominant source of these artifacts. 
This finding suggests that pose sensitivity, rather than view sparsity, fundamentally limits the robustness of splat-based CT. 
We therefore revisit the formulation of pose estimation in the splatting pipeline, examining how geometric perturbations propagate through differentiable projection operators and affect reconstruction stability. 
Our analysis leads to a refined gradient-based framework that allows joint optimization of volumetric and geometric parameters, ensuring stability against small pose deviations. 

In summary, this work provides both empirical and theoretical insights into pose sensitivity in splat-based CT and offers a practical self-calibrating formulation that mitigates artifacts under sparse-view conditions. 
Our main contributions are as follows:
\begin{itemize}
    \item We conduct the first systematic analysis of pose sensitivity in splat-based CT reconstruction, revealing that geometric misalignment, rather than data sparsity, is the dominant source of artifacts.
    \item We reformulate the gradient-based pose optimization in splatting, deriving a stable and differentiable calibration framework that jointly refines geometry and volume.
    \item We present an unbiased simulation framework for generating cone-beam datasets with controlled pose perturbations, enabling reproducible evaluation of geometric sensitivity. 
\end{itemize}

\section{Related Work}
\label{sec:related_work}
This section reviews recent advances in cone-beam CT reconstruction.

\mparagraph{Cone-Beam CT Reconstruction}
Cone-beam reconstruction algorithms in X-ray CT can be categorized by the way the volume is represented.
Voxel-based algorithms construct a cubic grid in 3D space and estimate the value of each element.
A representative method is the FDK algorithm~\cite{feldkamp1984practical}, which leverages the Fourier slice theorem, resulting in fast and reliable performance that has made it a standard algorithm.
Iterative reconstruction methods~\cite{sart_andersen,asd_pocs}, on the other hand, use a ray-projection model based on the Beer–Lambert law and iteratively optimize the voxel values to fit the model.
Advances in computer graphics and neural networks have enabled implicit representation-based reconstruction methods~\cite{naf_zha, neat2022, zang2021intratomo, sax_nerf}.
These approaches represent the volume using neural networks instead of explicit voxel grids. 
These methods perform well even under sparse-view conditions; however, due to the dense modeling of a volume using neural networks, their time complexity is generally significantly higher.
R$^2$-Gaussian~\cite{r2_gaussian} introduces a new approach to CT reconstruction using Gaussian splatting~\cite{gs_original}, where the volume is represented as a combination of 3D Gaussian primitives~\cite{ewa_splat}. 
This approach shows fast and high-quality reconstruction.

\mparagraph{Artifact Analysis}
CT artifacts arise in various forms depending on their sources and have been extensively studied~\cite{barrett2004artifacts}. For example, ring artifacts appear as circular bands and are typically caused by defective detector pixels~\cite{wedekind2023feature,croton2019ring,vsalplachta2021complete,liang2017iterative}. Metal artifacts arise from the polychromatic nature of X-rays~\cite{limar,nmar,polyner}. Sparse-view artifacts manifest as directional streaks due to insufficient projection data~\cite{kim2022streak,wang2022improving,lahiri2023sparse,loli2022comparison,champley2024methods}. Calibration artifacts appear as blurring or subtle streaks~\cite{thies2024gradient,thies2023gradient,gao2023fully,neat2022,thies2024differentiable,ruckert2024uncalibrated,wu2023joint}, originating from inaccuracies in system geometry.
Recently, neural networks have been actively used to classify artifacts and identify their causes~\cite{inkinen2024computed,prakash2019deep,madesta2024deep}.
However, works in this category take a slice from the reconstructed volume to produce heat maps of each artifact, so they do not account for these in 3D, and only partial 2D features from slices can be leveraged.
Additionally, the performance of these methods is significantly constrained due to the limited access to real artifact datasets.

\mparagraph{Geometric Calibration}
Precise calibration of a geometry configuration is crucial to the quality of the CT reconstruction.
This is a challenging problem, particularly in rotating systems, which can result in discrepancies between the actual and ideal geometries.
Calibration methods can be broadly categorized into~\textit{offline} and~\textit{online} approaches~\cite{geocalib_review}. The offline methods~\cite{offlinecalib_cbct, offlinecalib_dentalcbct, offlinecalib_ccbct} use specially manufactured phantoms of known size to estimate accurate geometry.
Although feature extraction from pre-scanned phantoms can yield precise calibration parameters, these methods cannot accommodate real-time geometric perturbations during scanning.
Moreover, they require additional scanning time and are primarily used for initial system setup. 
Online approaches~\cite{onlinecalib_meng,onlinecalib_2d3dreg,onlinecalib_ccbct}, in contrast, calibrate based on the actual target object.
The calibration parameters are jointly optimized during the reconstruction process, thereby improving quality and eliminating the need for an additional scan.
These approaches calculate a loss based on reconstruction quality and compute gradients with respect to camera pose parameters to iteratively reduce the loss by pose errors. 
Thies et al.~\cite{thies2024gradient} perform FDK-based reconstruction while correcting head motion, but the FDK limits the output quality and is dependent on a pre-trained network to measure the quality of reconstruction. 
Gao et al.~\cite{gao2023fully} reduce pose errors using a pre-trained projection network with segmented object volumes and captured projections, but their method requires ground-truth segmentation as prior information. 
Wu et al.~\cite{wu2023joint} optimize an implicit representation network while correcting camera poses, but their approach supports only the fan-beam geometry and does not provide publicly available code. 
R{\"u}ckert et al.~\cite{neat2022} employ an adaptive octree structure for an efficient implicit representation network with pose correction. %
In our method, we employ a splat-based method~\cite{r2_gaussian} that offers fast, robust, and fully differentiable reconstruction.
We further study the geometry parameters in~\cite{r2_gaussian} that support volumetric X-ray projections with various poses, which differ from those of the splatting method in the novel view synthesis~\cite{gs_original}.

\section{Analysis}
\label{sec:analysis}
\begin{figure}[htp]
	\centering
	\includegraphics[width=1.0\linewidth]{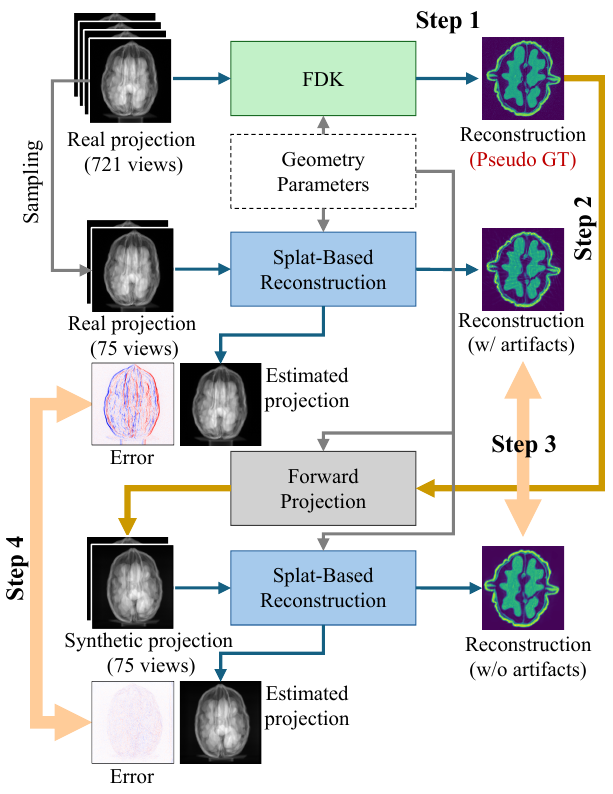}%
	\vspace{-1mm}
	\caption{\label{fig:walnut_artifact_analysis_pipeline}
		The pipeline of our artifact analysis experiment. 
		(Step~1) A pseudo ground truth (GT) volume is reconstructed using 721-view projections from the real dataset by the FDK~\cite{feldkamp1984practical}.
		(Step~2) We synthesize the 75 projection images from the pseudo-GT volume.
		(Step~3) We compare the results from the splat reconstruction~\cite{r2_gaussian} using both real and synthetic projections. 
		(Step~4) We compare the error of the estimated projections from the splat reconstruction between the real dataset and the synthesized dataset to identify the source of the error.}
	\vspace{-4mm}
\end{figure}

\begin{figure}[htp]
	\centering
	\includegraphics[width=0.6\linewidth]{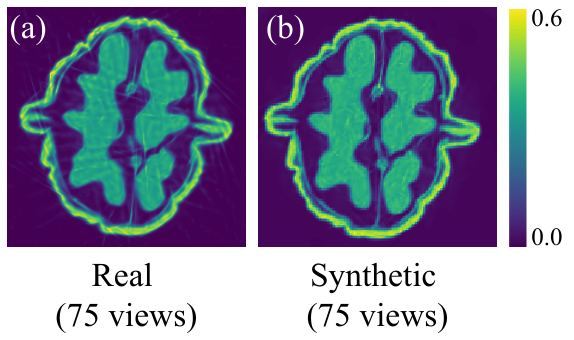}%
	\vspace{-2mm}
	\caption{\label{fig:walnut_synthetic_real_reconstruction}
		Comparison between the reconstructions from real 75-view projections and synthetic 75-view projections~(Figure~\ref{fig:walnut_artifact_analysis_pipeline}).
		The artifacts are observed only in the reconstruction from real projections, indicating the sparse input is not the main cause of the streaky artifacts.
	}
	\vspace{-4mm}
\end{figure}

\begin{figure}[htp]
	\centering
	\includegraphics[width=1.0\linewidth]{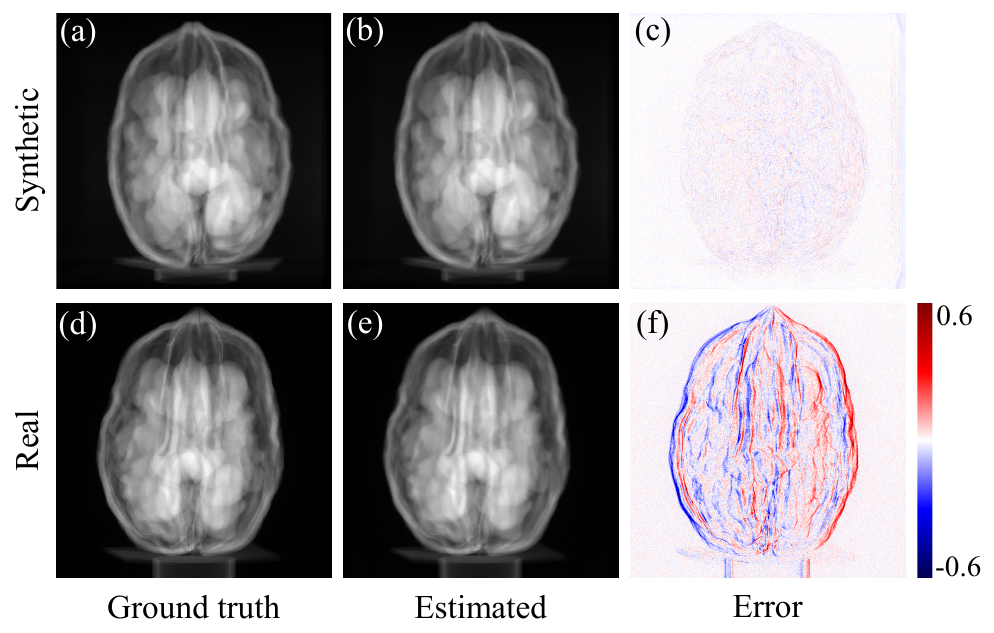}%
	\vspace{-2mm}
	\caption{\label{fig:walnut_proj_errors_red_blue}
		Comparison of projection images and error maps between the real and synthetic datasets with 75 projections.
		(a) and (b) show projections from the synthetic dataset and the estimated one from the splat reconstruction~\cite{r2_gaussian}.
		(d) and (e) show projections from the real dataset and the estimated one from the splat method.
		The difference between the ground truth and the estimated projections using the synthetic data is uniformly distributed (c); however, with the real data, it shows a~\textit{directional bias} around edges~(f), revealing the inaccuracy of camera poses.
	}
	\vspace{-3mm}
\end{figure}

In the splat baseline work~\cite{r2_gaussian}, the results from the real datasets in the sparse-view setting show streaky artifacts~(Figure~\ref{fig:walnut_synthetic_real_reconstruction}(a)), even though there were no metallic objects that induced anomalies in the scene.
We study this problem by designing experiments shown in Figure~\ref{fig:walnut_artifact_analysis_pipeline}.
We first obtain a pseudo ground-truth volume by reconstructing with FDK using dense 721 projections, under the assumption that FDK can produce a sufficiently accurate reconstruction when enough projections are available (Step 1 in Figure~\ref{fig:walnut_artifact_analysis_pipeline}). Generally, the root mean squared error (RMSE) scales as
\begin{equation}
	\mathrm{RMSE} \propto \frac{1}{\sqrt{N}},
\end{equation}
where $N$ denotes the number of projections under unbiased beam geometry errors. The mathematical derivation supporting this assumption is provided in the supplemental material.
Then, we create projection images from the reconstructed volume, which consists of 75 views, serving as a sparse-view set of projections.
We run the splat reconstruction~\cite{r2_gaussian} using the regenerated 75 projections to obtain another volume.
Notably, in the volume from the recreated projections, the streaky artifacts are significantly suppressed~(Figure~\ref{fig:walnut_synthetic_real_reconstruction}(b)) compared with those from the 75 projections from the original real projections~(Figure~\ref{fig:walnut_synthetic_real_reconstruction}(a)).
According to this experiment, we find that the streaky artifacts do not mainly come from the sparsity of input projections.
To further investigate the cause, we compare the error of the estimated projections from the splat reconstruction between the synthesized dataset and the real dataset~(Figure~\ref{fig:walnut_proj_errors_red_blue}). 
The difference between the ground truth and the estimated projections using the synthetic data is uniformly distributed~(Figure~\ref{fig:walnut_proj_errors_red_blue}(c)); however, with the real data, it shows a~\textit{directional bias} around edges~(Figure~\ref{fig:walnut_proj_errors_red_blue}(f)). 
From this observation, we conclude that the artifacts from the real dataset, when using the splat reconstruction, mainly originate from geometric errors.

\section{Method}
\label{sec:method}
\mparagraph{Projection Formulation}
We first restate the forward projection in the splatting formulation to examine how small pose perturbations affect the rendered intensity. 
Unlike the voxel-based Beer–Lambert model, the Gaussian representation introduces anisotropic weighting along the ray direction~\cite{r2_gaussian}, 
making the reconstructed intensity inherently sensitive to geometric misalignment. 
Formally, the attenuation field is modeled as:
\begin{equation}
	\label{eq:r2gs_atten}
	\begin{aligned}		
		\mu(\mathbf{x}) = \sum_{i}^{M} \rho_i \exp \left( -\frac{1}{2} (\mathbf{x} - \mathbf{p}_i)^{T} \mathbf{\Sigma}_i^{-1} (\mathbf{x} - \mathbf{p}_i) \right),
	\end{aligned}
\end{equation}
where $\mathbf{x}$ is the spatial query point, $M$ is the number of Gaussians, $\rho_i$ is the density of the $i$-th Gaussian, and~$\mathbf{p}_i\in \mathbb{R}^{3 \times 1}$,~$\mathbf{\Sigma}_i\in \mathbb{R}^{3 \times 3}$ denote its center and covariance, respectively. The forward projection of a ray is then computed by integrating the contribution of all Gaussians along the ray path:
\begin{equation}
	\label{eq:r2gs_proj}
	\scalebox{0.82}{$
	\begin{aligned}		
		I \left(\hat{\mathbf{x}}_{k}\right)
		&= \sum_{i}^M \sqrt{\dfrac{2\pi |\breve{\mathbf{\Sigma}}_{i,k}|}{|\hat{\mathbf{\Sigma}}_{i,k}|}} \, \rho_i \exp\left( -\frac{1}{2} (\hat{\mathbf{x}}_{k} - \hat{\mathbf{p}}_{i,k})^T \hat{\mathbf{\Sigma}}_{i,k}^{-1} (\hat{\mathbf{x}} - \hat{\mathbf{p}}_{i,k}) \right),
	\end{aligned}
	$}
\end{equation}
where~$\hat{\mathbf{x}}_{k}\in \mathbb{R}^{2 \times 1}$ is the projected query point,~$\breve{\mathbf{\Sigma}}_{i,k}\in \mathbb{R}^{3 \times 3}$ is the ray-space covariance,~$\hat{\mathbf{\Sigma}}_{i,k}\in \mathbb{R}^{2 \times 2}$ is the projected Gaussian covariance, and~$\hat{\mathbf{p}}_{i,k}\in \mathbb{R}^{2 \times 1}$ is the center coordinate of the projected Gaussian. $k$ is the index of the projection image. 
The parameters~$\rho_i, \mathbf{p}_i$, and~$\mathbf{\Sigma}_i$ are optimized to reduce the discrepancy between the computed projections from Equation~\eqref{eq:r2gs_proj} and the input projection images. 
After optimization, the final attenuation volume is obtained by summing up the Gaussian components at query points such as voxel grids. 

\mparagraph{Pose Parameterization}
To ensure stable optimization near the nominal geometry, we parameterize pose perturbations as small increments relative to the initial configuration. 
This incremental formulation mitigates gradient explosion under small-angle approximation and allows fine-grained refinement of rotation and translation during joint optimization.
We express the rigid body motion of each camera using a rotation matrix~$\mathbf{W}_k \in \mathbb{R}^{3 \times 3}$ and a translation vector~$\mathbf{t}_k \in \mathbb{R}^{3 \times 1}$, then the $i$-th Gaussian primitive~(Equation~\eqref{eq:r2gs_atten}) is transformed as:
\begin{equation}
	\label{eq:r2gs_view_transform}
	\scalebox{1.0}{$
		\tilde{\mathbf{p}}_{i,k}      = \mathbf{W}_k \mathbf{p}_i + \mathbf{t}_k,
		\quad
		\tilde{\mathbf{\Sigma}}_{i,k} = \mathbf{W}_k \mathbf{\Sigma}_i \mathbf{W}_k^{T},
		$}
\end{equation}
where~$\tilde{\mathbf{p}}_{i,k}$ and~$\tilde{\mathbf{\Sigma}}_{i,k}$ are the transformed Gaussian's coordinates and covariance matrix in 3D.
The transformed Gaussian is then projected onto the 2D detector as below:
\begin{equation}
	\label{eq:r2gs_proj_transform}
	\scalebox{1.0}{$
		\begin{aligned}
			\hat{\mathbf{p}}_{i,k}      = \psi\left(\tilde{\mathbf{p}}_{i,k}\right), 
			\;
			\hat{\mathbf{\Sigma}}_{i,k} = \left( \mathbf{J}_{i,k} \tilde{\mathbf{\Sigma}}_{i,k} \mathbf{J}_{i,k}^{T} \right)_{\{0,1\},\{0,1\}},
		\end{aligned}
		$}
\end{equation}
where $\psi\left(\cdot\right):\mathbb{R}^{3 \times 1}\rightarrow\mathbb{R}^{2 \times 1}$ is the non-linear projection function,~${\left(  \cdot  \right)_{\left\{ {0,1} \right\},\left\{ {0,1} \right\}}}$ is an operator that crops the first~$2\times2$ submatrix from a matrix, and $\mathbf{J}_{i,k}$ is the Jacobian of projection function with respect to the camera-space position:
\begin{equation}
	\label{eq:r2gs_jacov}
	\scalebox{1.0}{$
		\begin{aligned}		
			\mathbf{J}_{i,k} &= \dfrac{\partial \phi\left(\tilde{\mathbf{p}}_{i,k}\right)}{\partial \tilde{\mathbf{p}}_{i,k}} \in \mathbb{R}^{3 \times 3}, \\[0.3em]
			\phi\left(\tilde{\mathbf{p}}_{i,k}\right) &=
			\begin{pmatrix}
				\hat{\mathbf{p}}_{i,k} \\
				\|\tilde{\mathbf{p}}_{i,k}\|
			\end{pmatrix} \in \mathbb{R}^{3 \times 1}
		\end{aligned}
		$}
\end{equation}
where~$\phi\left(\cdot\right)$ is the mapping function from the camera-space to ray-space used in the volumetric EWA splatting~\cite{ewa_splat}.
Note that in the original Gaussian Splatting~\cite{gs_original}, this mapping function does not include the last component, so it becomes 2D.
On the other hand, in the CT splatting the third component in~$\phi\left(\cdot\right)$ is included, since we need 3D information of the mapping function to compute the ray integral of the Gaussian primitives located in the real distance to simulate the X-ray attenuation energy.

To optimize the pose parameters, we model the motion using the 4D quaternion for rotation and the 3D vector for translation, following prior works~\cite{liugs,zhao2024sparseags,scgs_deng,scgs_shi}.
Unlike conventional CT calibration methods, our approach optimizes these parameters simultaneously with the volume, allowing for self-calibrating reconstruction. 
To ensure fine-grained optimization around the initial geometric configuration, we parameterize pose errors relative to the prior camera pose by introducing a motion increment as follows~\cite{scgs_deng}:
\begin{equation}
	\label{eq:param_quat_tran}
	\scalebox{0.95}{$
		\Delta \mathbf{q}_k = \mathbf{q}_k - \mathbf{q}_{k,\mathrm{init}} \in \mathbb{R}^{4 \times 1},
		\quad
		\Delta \mathbf{t}_k = \mathbf{t}_k - \mathbf{t}_{k,\mathrm{init}} \in \mathbb{R}^{3 \times 1},
		$}
\end{equation}
where $\mathbf{q}_k$ and $\mathbf{t}_k$ represent the quaternion and translation vectors, while $\Delta \mathbf{q}_k$ and $\Delta \mathbf{t}_k$ denote the small pose correction terms optimized during reconstruction. 
This formulation models a small linearized increment in motion, allowing more accurate pose refinement.
The quaternion and translation parameters are converted to the rotation matrix and the translation vector:
\begin{equation}
	\label{eq:quat_tran_to_view}
	\scalebox{1.0}{$
		\begin{aligned}
			\mathbf{W}_k = \mathrm{Rot}(\Delta \mathbf{q}_k + \mathbf{q}_{k,\mathrm{init}}),
			\quad
			\mathbf{t}_k = \Delta \mathbf{t}_k + \mathbf{t}_{k,\mathrm{init}},
		\end{aligned}
		$}
\end{equation}
where $\mathrm{Rot}\left(\cdot\right)$ is the rotation mapping function from the quaternion vector.
The parameters for our joint calibration and reconstruction method are as follows:
\begin{equation}
	\label{eq:opt_params}
	\scalebox{1.0}{$
		\mathbf{\Theta} = \{\mathbf{p}_i, \mathbf{\Sigma}_i, \rho_i, \Delta\mathbf{q}_k, \Delta\mathbf{t}_k\}.
		$}
\end{equation}

\mparagraph{Optimization}
Our optimization problem is given by:
\begin{equation}
	\label{eq:opt_argmin}
	\scalebox{0.9}{$
		\mathbf{\Theta} = 
		\operatorname*{arg\,min}_{\mathbf{\Theta}}\mathcal{L}(I_k, \hat{I}_k),
		$}
\end{equation}	
where $\mathcal{L}$ is the loss function, and~$I_k, \hat{I}_k \in \mathbb{R}^{H \times W}$ denote the given and estimated projection images respectively.
Our loss function is defined as:
\begin{equation}
	\label{eq:loss_total}
	\scalebox{1.0}{$
		\mathcal{L}(I_k, \hat{I}_k) = \mathcal{L}_{L1}(I_k, \hat{I}_k) + \lambda \mathcal{L}_{\mathrm{SSIM}}(I_k, \hat{I}_k).
		$}
\end{equation}
The objective function combines pixel-wise and structural similarity losses without total variation (TV) regularization. 
We find that, under our joint calibration framework, the explicit TV term becomes unnecessary and can even suppress essential geometric gradients, 
thereby reducing the system's ability to recover fine pose corrections. 
This observation empirically supports that stability arises naturally from accurate geometric modeling rather than from heuristic smoothness constraints.

\begin{figure}
	\centering
	\includegraphics[width=1.0\linewidth]{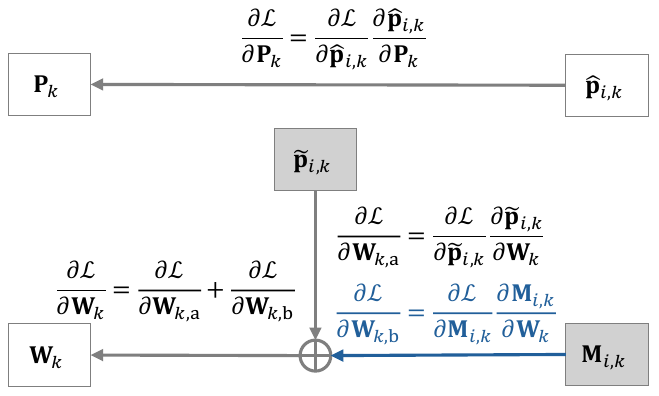}%
	\vspace{-3mm}
	\caption{\label{fig:call_graph}
		The backward gradient computation pipeline with respect to the intermediate values $\mathbf{P}_k$ and $\mathbf{W}_k$, which are directly dependent on the camera pose parameters.
		$\mathbf{P}_k \in \mathbb{R}^{3\times4}$ is the perspective projection matrix that transforms 4D homogenous coordinates to the 3D ones.
		$\mathbf{M}_{i,k} \in \mathbb{R}^{3\times3}$ is the multiplication of $\mathbf{J}_{i,k}$ and $\mathbf{W}_{k}$. 
		Note that the term $\frac{\partial\mathcal{L}}{\partial\mathbf{W}_{k,\text{b}}}$ in blue is different from the calibration method in the RGB splatting~\cite{scgs_deng}.
	}
	\vspace{-2mm}
\end{figure}

Figure~\ref{fig:call_graph} illustrates the backward gradient flow through pose-dependent transformations. 
We explicitly track the Jacobian terms with respect to both the camera projection matrix $\mathbf{P}_k$ and the rotation matrix $\mathbf{W}_k$, 
allowing us to observe how small geometric perturbations propagate through the differentiable rendering pipeline. 
This analysis provides insight into the source of instability in prior splat-based CT methods, which often neglect these cross-derivative dependencies.

\section{Results}
\subsection{Experiment Setup}
We implement our method using PyTorch~\cite{pytorch} and CUDA~\cite{cuda}, running experiments on an Intel Xeon 4214R CPU (2.4 GHz) and an NVIDIA RTX A6000 GPU. 
For optimization, we adopt the Gaussian learning rates from R$^2$-Gaussian~\cite{r2_gaussian} while setting the learning rate for camera parameters to 2e-4, which exponentially decays to 2e-5 over 30,000 steps.
For synthetic data generation, we use TIGRE-v2.3~\cite{tigre_toolbox}, while real-world evaluations are conducted on the CT dataset from~\cite{real_dataset}. 
This setup ensures a controlled evaluation on synthetic data and a realistic assessment of our method's robustness in real-world scenarios.

\subsection{Dataset Generation with Geometric Error}
To validate our self-calibration framework, we generate a CT dataset with simulated geometric errors. 
Conventional synthetic datasets typically assume an ideal circular trajectory, where the sensor rotates around a fixed center of rotation. 
However, mechanical imperfections in real systems introduce deviations from this ideal configuration, resulting in misaligned projections and degraded reconstruction quality. 
To simulate these real-world effects, we apply perturbations to the camera parameters during projection generation, embedding mechanical misalignment into the dataset.
We model the detector geometry as a rigid-body transformation $\mathbf{T}$ in the special Euclidean group $\mathrm{SE}(3)$. 
To introduce realistic and unbiased perturbations, we treat rotation and translation errors separately. Since translation $\mathbf{t}$ is linear, we sample it from a zero-mean Gaussian distribution:
$\mathbf{t}\sim\mathcal{N}(0, \sigma_\text{trans}^2\mathbf{I}_{3\times3})$,
where $\sigma_\text{trans}^2$ controls the magnitude of translation variation. 
In contrast, rotation $\mathbf{R}$ lies in the nonlinear space $\mathrm{SO}(3)$, preventing direct additive perturbations. 
Instead, we apply a logarithmic map to transform it into the tangent space, add Gaussian noise with variance $\sigma_\text{rot}^2$, and then apply the exponential map to project it back to $\mathrm{SO}(3)$~\cite{sola2018micro}. 
This ensures realistic rotational perturbations while preserving valid transformations. 
Note that because translations are sampled in voxel space, their physical effect may vary with dataset scaling.
By incorporating these calibrated perturbations, our dataset provides an unbiased and realistic testbed for evaluating self-calibration in X-ray CT reconstruction. 
Mathematical proofs of the unbiasedness of our dataset generation are provided in the supplemental material.

\begin{figure*}[t]
	\centering
	\includegraphics[width=0.8\linewidth]{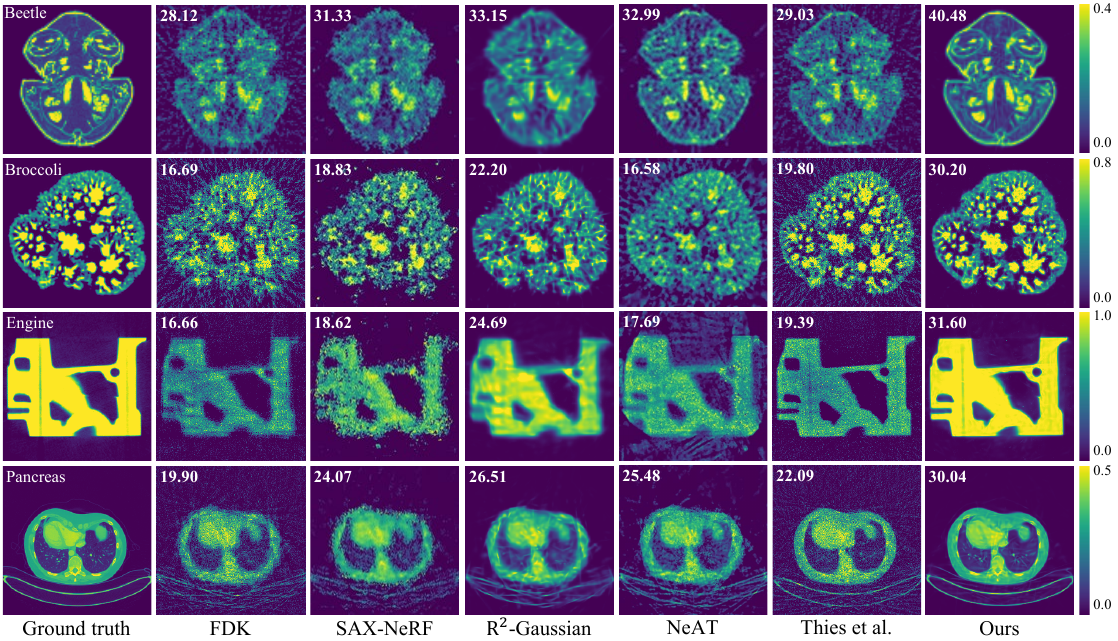}%
	\vspace{-2mm}
	\caption{\label{fig:results_synthetic_comparison}
		Comparison of reconstructed volume slices on the synthetic dataset with the pose perturbation. We compare our method against existing approaches, including FDK~\cite{feldkamp1984practical}, SAX-NeRF~\cite{sax_nerf}, R$^2$-Gaussian~\cite{r2_gaussian}, NeAT~\cite{neat2022}, and Thies et al.~\cite{thies2024gradient} across multiple objects. The numbers in the upper-left corners of each image indicate the PSNR values (dB) relative to the respective ground truth volume. Our method achieves higher PSNR values and better visual quality, demonstrating superior robustness against geometric misalignments.}
	\vspace{-2mm}
\end{figure*}

\begin{figure*}[t]
	\centering
	\includegraphics[width=0.8\linewidth]{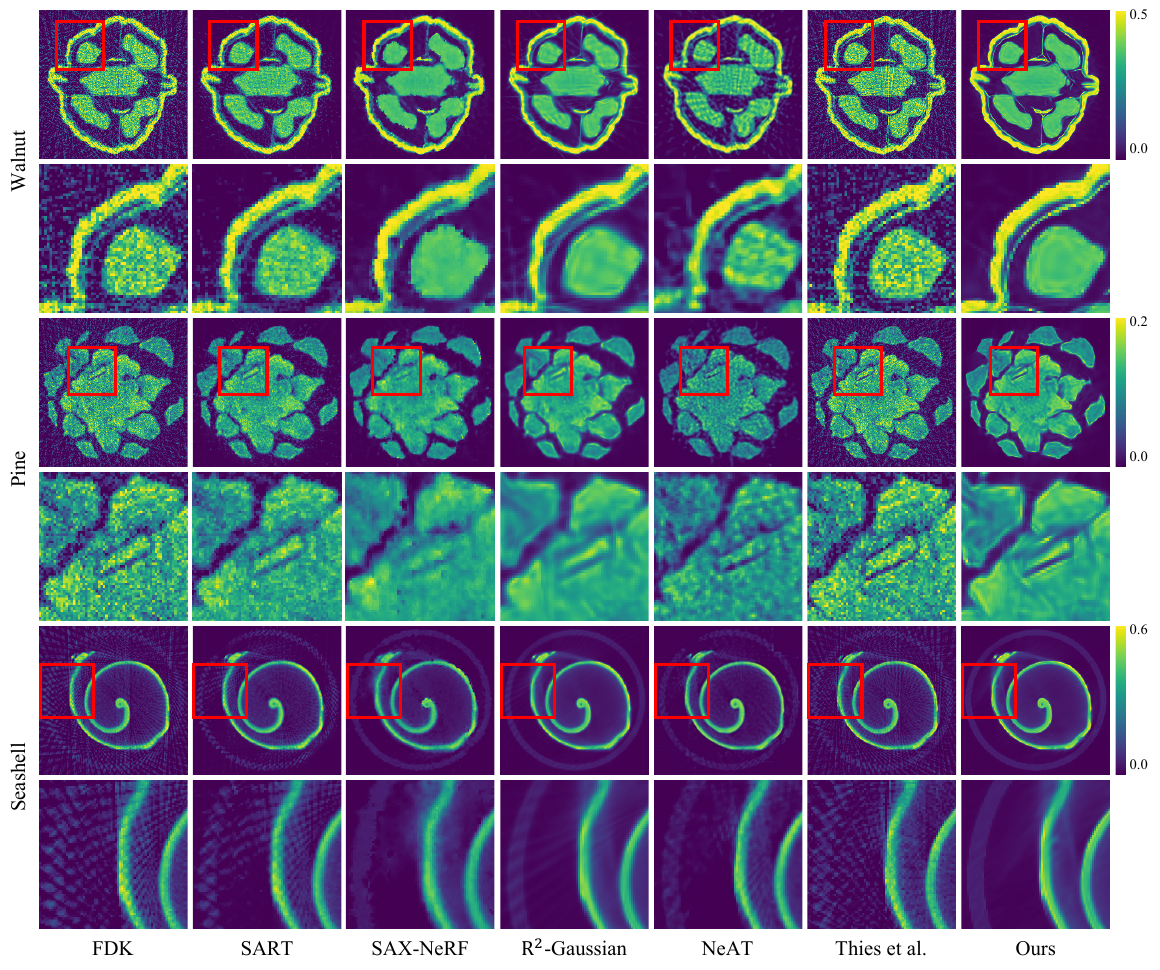}%
	\vspace{-2mm}
	\caption{\label{fig:results_real_comparison}
		Comparison of reconstructed volume slices on the real dataset~\cite{real_dataset}. All methods reconstruct the volume using 75 projections. Our approach achieves superior performance, effectively mitigating reconstruction artifacts and preserving fine details, thanks to our self-calibrating geometric correction.
		The zoomed-in regions (highlighted in red boxes) demonstrate that our method recovers finer structures with fewer artifacts compared to other approaches.}	
	\vspace{0mm}
\end{figure*}

\subsection{Evaluation}

\mparagraph{Quantitative Evaluation}
We evaluate our method on synthetic datasets and present the reconstructed volume results in Figure~\ref{fig:results_synthetic_comparison}. 
We compare our approach against traditional tomography methods~\cite{feldkamp1984practical}, sparse-view reconstruction methods~\cite{sax_nerf,r2_gaussian}, and the state-of-the-art joint calibration and reconstruction methods~\cite{neat2022,thies2024gradient}.
To simulate real-world geometric miscalibrations, we introduce camera rotation and translation noise with standard deviations of 0.03 in the Lie algebra domain and 1.0 in unit length (corresponding to the voxel size of the reconstruction volume), respectively.
Our method consistently outperforms all baseline methods both quantitatively and qualitatively, demonstrating robustness under geometric perturbations. 
Note that our approach exceeds the SOTA method~\cite{thies2024gradient} by approximately 10 dB in PSNR, highlighting its superior reconstruction accuracy. 
In contrast, existing methods often fail to recover accurate volumes under geometric errors, as they rely on a pre-calibrated hardware setup or lack of self-correction performance. 
To further validate the effectiveness of our approach, we report quantitative results across different synthetic datasets, with and without geometric perturbations on the baseline splat-based reconstruction method~\cite{r2_gaussian} and ours (Table~\ref{tab:results_with_without_noise_psnr_ssim}).
On the noise-free dataset, the baseline and ours indicates similar performance. But on the noisy dataset, ours shows significantly better results compared to the baseline.
These results confirm that our method effectively reconstructs volumes while simultaneously handling camera pose errors. 
\begin{table}[h]
	\centering
	\caption{\label{tab:results_with_without_noise_psnr_ssim}
		Comparison of the baseline splat-based reconstruction~\cite{r2_gaussian} and ours on the noise-free and noisy datasets.
		Values are reported in PSNR.}
	\resizebox{0.7\linewidth}{!}{
	\begin{tabular}{lcccc}
		\toprule
		\multirow{2}{*}{Scene} & \multicolumn{2}{c}{Noise-free dataset} & \multicolumn{2}{c}{Noisy dataset} \\
		\cmidrule(lr){2-3} \cmidrule(lr){4-5}
		& Baseline & Ours & Baseline & Ours \\
		\midrule
		Chest     & 35.81 & 35.68 & 26.69 & 30.44 \\
		Foot      & 32.51 & 32.04 & 25.46 & 30.57 \\
		Head      & 40.99 & 40.92 & 30.91 & 34.84 \\
		Jaw       & 37.27 & 36.77 & 27.12 & 32.38 \\
		Pancreas  & 37.91 & 37.50 & 26.51 & 30.04 \\
		Beetle    & 43.18 & 43.22 & 33.15 & 40.48 \\
		Bonsai    & 35.62 & 34.99 & 26.61 & 30.05 \\
		Broccoli  & 36.54 & 34.70 & 22.21 & 30.20 \\
		Kingsnake & 39.85 & 39.97 & 35.86 & 38.16 \\
		Pepper    & 39.40 & 38.68 & 26.51 & 31.96 \\
		Backpack  & 38.69 & 38.90 & 27.88 & 32.19 \\
		Engine    & 40.25 & 39.33 & 24.69 & 31.60 \\
		Mount     & 38.94 & 38.08 & 29.07 & 31.13 \\
		Present   & 38.06 & 37.82 & 28.20 & 33.84 \\
		Teapot    & 47.81 & 47.79 & 36.65 & 43.43 \\
		\bottomrule
	\end{tabular}
	}
	\vspace{-3mm}
\end{table}
In addition to reconstruction accuracy, we evaluate geometric calibration performance in Table~\ref{tab:results_calibration_error}.
We measure camera pose estimation errors using the transformation:
\begin{align} 
	\mathbf{T}=\begin{bmatrix} \mathbf{R} & \mathbf{t} \\ \mathbf{0}_3^t & 1\end{bmatrix}=\mathbf{T}_\text{gt}^{-1} \mathbf{T}_\text{est} ,
\end{align} 
where $\mathbf{T}_\text{gt}$ and $\mathbf{T}_\text{est}$ represent the ground truth and estimated camera transformations, respectively. We compute the orientation error as
$\theta = \frac{1}{N}\sqrt{\arccos{\frac{\mathrm{tr}(\mathbf{R}) - 1}{2}}}$,
and define the translation error as the Euclidean distance between the estimated and ground truth camera positions. Table~\ref{tab:results_calibration_error} reports the root mean squared error (RMSE) of $\theta$ and $\mathbf{t}$, demonstrating that our self-calibration framework significantly reduces pose estimation errors.
These results highlight the effectiveness of our method in simultaneously optimizing volume reconstruction and geometric calibration, making it a robust solution for real-world X-ray tomography applications.

\begin{table}[pt]
	\centering
	\caption{\label{tab:results_calibration_error}
		We report the root mean squared error~(RMSE) of translation and orientation errors after applying our self-calibration framework. 
		The translation error is expressed in arbitrary units (1 AU corresponds to one voxel size of the reconstruction volume), and the orientation error is measured in degrees. 
		Our method outperforms over the state-of-the-art online calibrating reconstruction methods with significantly reduced pose errors.
		}\vspace{-2mm}
	\resizebox{1.0\linewidth}{!}{
		\begin{tabular}{lcccccc}
			\toprule
			Scene & \multicolumn{3}{c}{Translation (AU)} & \multicolumn{3}{c}{Orientation (\textdegree)} \\
			\cmidrule(lr){2-4} \cmidrule(lr){5-7}
			& NeAT~\cite{neat2022} & Thies et al.~\cite{thies2024gradient} & Ours & NeAT~\cite{neat2022} & Thies et al.~\cite{thies2024gradient} & Ours \\
			\midrule
			Chest     & 1.076 & 2.347 & \textbf{0.973} & 2.720 & 3.919 & \textbf{0.621} \\
			Foot      & 1.768 & 2.685 & \textbf{1.213} & 2.867 & 4.434 & \textbf{0.461} \\ 
			Head      & 1.254 & 2.371 & \textbf{0.726} & 3.019 & 3.682 & \textbf{0.782} \\
			Jaw       & 1.310 & 2.616 & \textbf{0.924} & 2.872 & 4.295 & \textbf{0.523} \\
			Pancreas  & 1.455 & 2.587 & \textbf{1.247} & 3.058 & 3.649 & \textbf{1.020} \\
			Beetle    & 1.076 & 2.037 & \textbf{0.809} & 3.030 & 3.144 & \textbf{0.194} \\
			Bonsai    & 2.804 & 3.083 & \textbf{0.732} & 3.035 & 3.961 & \textbf{0.747} \\
			Broccoli  & 1.475 & 2.095 & \textbf{0.671} & 2.857 & 5.580 & \textbf{0.314} \\
			Kingsnake & 1.097 & 2.025 & \textbf{0.484} & 2.983 & 3.214 & \textbf{1.590} \\
			Pepper    & 1.469 & 2.711 & \textbf{0.397} & 2.789 & 4.892 & \textbf{0.629} \\
			Backpack  & 1.570 & 2.413 & \textbf{0.549} & 2.894 & 3.412 & \textbf{0.563} \\
			Engine    & 1.632 & 2.357 & \textbf{0.598} & 2.739 & 5.845 & \textbf{0.484} \\
			Mount     & 1.169 & 2.768 & \textbf{0.769} & 2.550 & 4.313 & \textbf{0.567} \\
			Present   & 1.243 & 2.546 & \textbf{0.399} & 3.022 & 3.784 & \textbf{0.577} \\
			Teapot    & 1.152 & 2.308 & \textbf{0.397} & 2.781 & 3.017 & \textbf{0.339} \\
			\midrule
			Mean      & 1.437 & 2.463 & \textbf{0.726} & 2.881 & 4.076 & \textbf{0.627} \\
			\bottomrule
		\end{tabular}

	}
	\vspace{-3mm}
\end{table}

\begin{table}[pt]
	\centering
	\caption{\label{tab:results_noise_level_test}%
		Reconstruction performance with camera noise varies. We compare the baseline~\cite{r2_gaussian} and our method under different camera pose perturbations. Standard deviations $\sigma_{\text{rot}}$ and $\sigma_{\text{trans}}$ indicate rotational and translational noise levels. Both methods perform well without noise (first row), but our method shows greater robustness to pose errors, achieving higher PSNR and SSIM in most scenario levels. The best scores for each experiment are highlighted in bold.}
	\vspace{-2mm}	
	\resizebox{0.97\linewidth}{!}{
		\begin{tabular}{c|cc|cc|cc}  
			\toprule
			\multirow{2}{*}{Scene} & \multirow{2}{*}{$\sigma_{\text{rot}}$} & \multirow{2}{*}{$\sigma_{\text{trans}}$} & \multicolumn{2}{c|}{Baseline~\cite{r2_gaussian}} & \multicolumn{2}{c}{Ours} \\
			\cmidrule{4-7}
			& & & PSNR (dB) & SSIM & PSNR (dB) & SSIM \\
			\midrule
			\multirow{5}{*}{Beetle} 
			& - & - & {43.18} & {0.995} & \textbf{43.22} & \textbf{0.995} \\			
			& 0.01 & 0.5 & {37.28} & {0.979} & \textbf{41.38} & \textbf{0.992} \\
			& 0.02 & 1.0 & {33.90} & {0.954} & \textbf{40.65} & \textbf{0.990} \\			
			& 0.05 & 2.5 & {31.91} & {0.930} & \textbf{35.42} & \textbf{0.970} \\						
			& 0.10 & 5.0 & {30.80} & {0.920} & \textbf{32.32} & \textbf{0.939} \\
			\midrule
			\multirow{5}{*}{Head} 
			& - & - & \textbf{40.99} & \textbf{0.983} & {40.92} & {0.981} \\
			& 0.01 & 0.5 & {35.29} & {0.932} & \textbf{37.87} & \textbf{0.954} \\
			& 0.02 & 1.0 & {32.25} & {0.883} & \textbf{35.63} & \textbf{0.944} \\			
			& 0.05 & 2.5 & {28.53} & {0.808} & \textbf{32.61} & \textbf{0.892} \\						
			& 0.10 & 5.0 & {26.41} & {0.768} & \textbf{28.17} & \textbf{0.805} \\
			\bottomrule
		\end{tabular}
	}
\end{table}

\mparagraph{Qualitative Evaluation}
Figure~\ref{fig:results_real_comparison} presents the reconstruction results on the real dataset~\cite{real_dataset} using 75 views.
The FDK reconstruction exhibits strong radial artifacts, which arise due to sparse-view sampling. 
While SART provides improved reconstruction quality over FDK, it still suffers from similar structured artifacts.
SAX-NeRF~\cite{sax_nerf} effectively suppresses the radial artifacts, but details are smoothed out due to calibration errors.
R$^2$-Gaussian~\cite{r2_gaussian} reduces anomalies but remains affected by needle-like artifacts, a consequence of imperfect camera calibration. 
To mitigate these issues, R$^2$-Gaussian employs TV regularization, which suppresses artifacts but over-smooths surfaces, as observed in the fourth column of Figure~\ref{fig:results_real_comparison}.
NeAT~\cite{neat2022} produces sharp object boundaries but fails to reconstruct homogeneous regions effectively.
Thies et al.~\cite{thies2024gradient} show slightly better performance than their baseline method~\cite{feldkamp1984practical}, but suffer from severe sparse-view artifacts inherited from the baseline FDK properties.
In contrast, our method achieves high-accuracy reconstruction, effectively recovering fine details while suppressing radial artifacts without relying on aggressive smoothing. 
By jointly optimizing volumetric reconstruction and geometric calibration, our approach ensures artifact-free and detail-preserving reconstructions, demonstrating clear advantages over existing methods.

\mparagraph{Computation Time}
To evaluate the computational efficiency of our method, we compare the average reconstruction time over all synthetic scenes across different approaches, as summarized in Table~\ref{tab:results_mean_computation_time}.
Our method achieves the lowest computation time among joint reconstruction and calibration methods, even outperforming the baseline~\cite{r2_gaussian}.
The reduction in computation time compared to the baseline mainly arises from the absence of the total variation (TV) regularization term in our formulation.
\begin{table}[tp]
		\vspace{-3mm}
	\centering
	\caption{Comparison of mean computation time (in minutes) among joint reconstruction and calibration methods, including the baseline.}
	\vspace{-3mm}
	\scalebox{0.8}{
		\begin{tabular}{cccc}
			\toprule
			\multicolumn{4}{c}{Computation time (min)} \\ 
			\cmidrule(lr){1-4}
			Baseline~\cite{r2_gaussian} & NeAT~\cite{neat2022} & Thies et al.~\cite{thies2024gradient} & Ours \\ 
			\midrule
			23.19 & 48.35 & 31.15 & \textbf{20.89} \\
			\bottomrule
		\end{tabular}
	}
	\vspace{-7mm}
	\label{tab:results_mean_computation_time}
\end{table}

\subsection{Ablation Studies}
We evaluate the reconstruction performance of our method against the baseline splat-based algorithm~\cite{r2_gaussian} under different levels of pose perturbation and varying numbers of input views.
As shown in Table~\ref{tab:results_noise_level_test}, both methods achieve high reconstruction accuracy in the absence of pose errors. 
However, as the perturbation level increases, the baseline performance rapidly deteriorates, while our method maintains stable results thanks to its self-calibrating design.
Table~\ref{tab:results_psnr_ssim} summarizes the results under pose perturbations ($\sigma_{\text{rot}}=0.03$, $\sigma_{\text{trans}}=1.0$) across 75, 50, and 25 views. 
As the number of views decreases, reconstruction quality drops for both methods, yet our approach consistently outperforms the baseline.
	These results demonstrate the robustness of our joint calibration module in mitigating geometric misalignment while preserving fine structural details.

\begin{table}[h]
	\centering
	\caption{\label{tab:results_psnr_ssim}
		Quantitative comparison under geometric perturbations with different numbers of input views. 
		Our method consistently outperforms the baseline~\cite{r2_gaussian} across 75, 50, and 25 views on synthetic datasets.}
	\vspace{-3mm}		
	\resizebox{0.55\linewidth}{!}{
		\begin{tabular}{ccccc}
			\toprule
			Views & \multicolumn{2}{c}{Baseline~\cite{r2_gaussian}} & \multicolumn{2}{c}{Ours} \\
			\cmidrule(lr){2-3} \cmidrule(lr){4-5}
			& PSNR & SSIM & PSNR & SSIM \\
			\midrule
			75 & 28.50 & 0.820 & \textbf{33.42} & \textbf{0.906} \\
			50 & 27.67 & 0.799 & \textbf{31.73} & \textbf{0.879} \\
			25 & 26.44 & 0.764 & \textbf{29.02} & \textbf{0.810} \\
			\bottomrule
		\end{tabular}
	}
	\vspace{-5mm}
\end{table}

\section{Limitation and Discussion}
\label{sec:discussion}

\begin{figure}[h]
	\centering
	\includegraphics[width=0.6\linewidth]{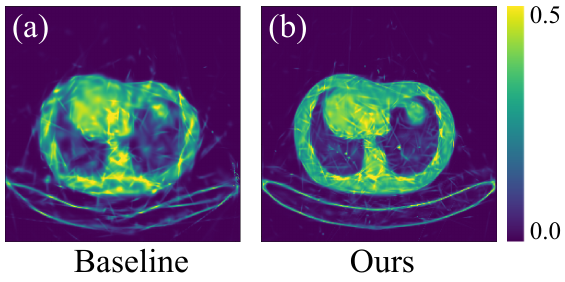}%
	\vspace{-2mm}
	\caption{\label{fig:limitation}
		Comparison between the baseline~\cite{r2_gaussian} and ours from extremely sparse input projections (25 views) with geometric perturbations.
		Our method reconstructs the volume with significantly fewer artifacts even under extreme conditions.
	}
	\vspace{-2mm}
\end{figure}

We further evaluate our method and the baseline method~\cite{r2_gaussian} under extremely sparse-view conditions (25 projections) with pose errors.
Compared with the baseline~(Figure~\ref{fig:limitation}(a)), our method still shows significantly fewer artifacts with the extremely sparse inputs~(Figure~\ref{fig:limitation}(b)). However, ours starts to contain the needle-like artifacts due to the absence of the TV regularization.
This suggests that while our self-calibration framework consistently mitigates geometric misalignment, additional regularization strategies may be necessary for extreme sparse-view scenarios. 
Addressing this limitation to enhance reconstruction quality in the extreme sparse inputs is future work.

\section{Conclusion}
We have presented an analysis method for localizing the root cause of stripe artifacts and a self-calibrating splat-based tomography framework that jointly optimizes volume reconstruction and geometric calibration. 
By introducing a differentiable loss with respect to camera parameters, our method corrects geometric errors and real-world misalignment in X-ray CT. 
Experiments demonstrate consistent reconstruction accuracy in the presence of hardware-induced geometric faults, outperforming existing approaches. 
Our method is robust to pose errors and does not require pre-calibrated hardware. 
Furthermore, the estimated calibration parameters can be reused in other tomographic reconstruction methods.

\appendix
\vspace{-0.1cm}
\section*{Acknowledgements}
\vspace{-0.15cm}
\noindent Min H.~Kim acknowledges the Samsung Research Funding \& Incubation Center (SRFC-IT2402-02), the Korea NRF grant (RS-2024-00357548), the MSIT/IITP of Korea (RS-2022-00155620, RS-2024-00398830, RS-2024-00436680, and 2017-0-00072), the MSIT Advanced GPU Utilization Support Program, and Microsoft Research Asia, with additional support by the KAIST Jang Young Sil Fellow Program.

\clearpage
{
    \small
    \bibliographystyle{ieeenat_fullname}
    \bibliography{bibliography}
}

\clearpage 
\pagestyle{empty} 
\includepdf[pages=-]{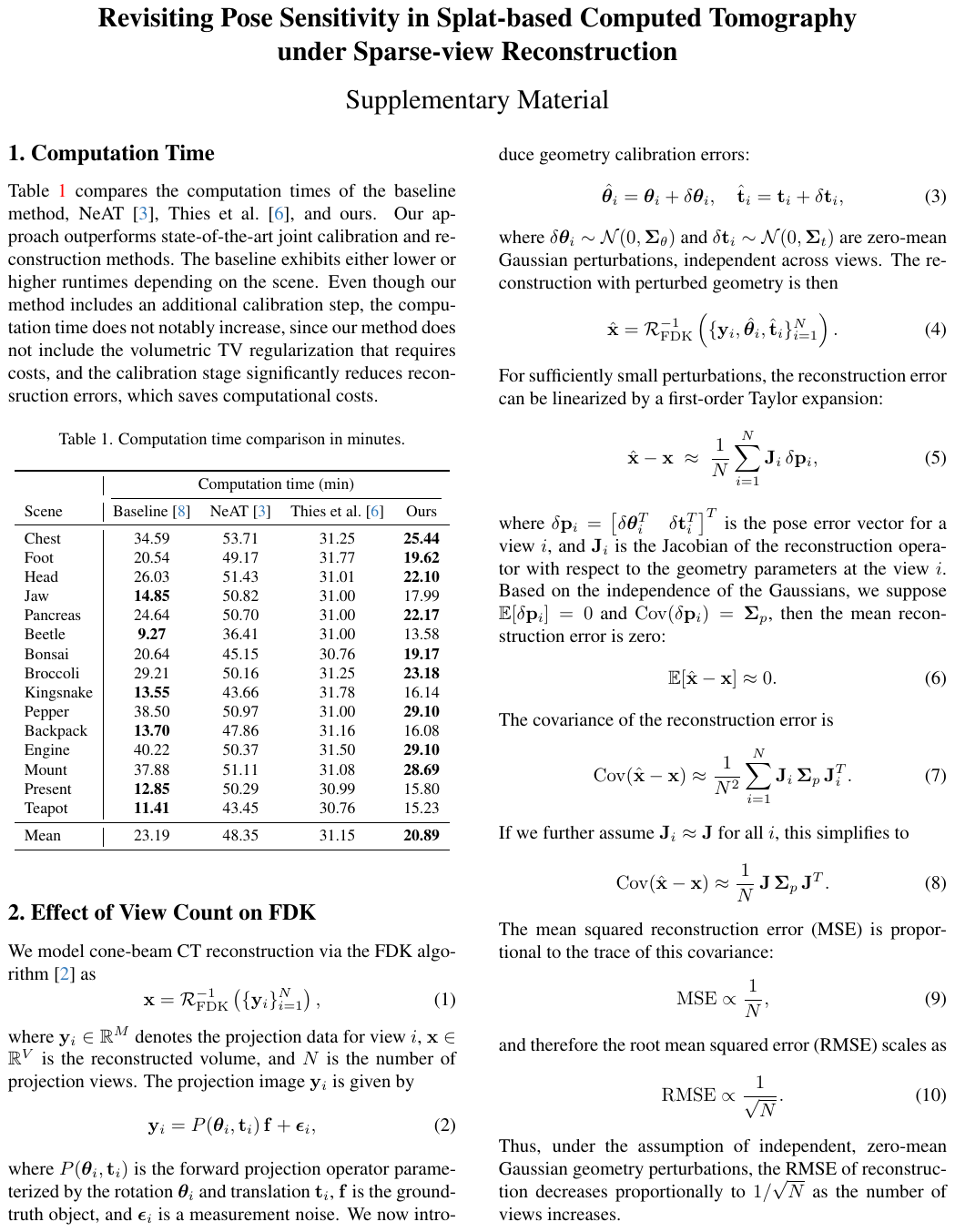}
\clearpage 
\pagestyle{plain} 

\end{document}